\documentclass[10pt,twocolumn,letterpaper]{article}

\usepackage[pagenumbers]{cvpr} %

\usepackage[utf8]{inputenc} %
\usepackage[T1]{fontenc}    %
\usepackage{booktabs}       %
\usepackage{amsfonts}       %
\usepackage{nicefrac}       %
\usepackage{microtype}      %
\usepackage{float}
\usepackage{colortbl}
\usepackage{graphicx}
\usepackage{dsfont}
\usepackage{amsmath,amssymb}
\usepackage{makecell}
\usepackage{wrapfig}
\usepackage{multirow}
\usepackage{pifont}
\usepackage{lipsum}
\usepackage{rotating}
\usepackage{bm}
\usepackage{adjustbox}
\usepackage{caption}
\usepackage{amsmath}
\usepackage{siunitx}
\usepackage{algorithm}
\usepackage{algpseudocode}
\usepackage[utf8]{inputenc}      %
\usepackage[T1]{fontenc}         %
\usepackage{amsmath,amssymb}     %
\usepackage{algorithmicx}    %
\usepackage{url}            %
\usepackage[dvipsnames]{xcolor}

\definecolor{cvprblue}{rgb}{0.21,0.49,0.74}
\usepackage[pagebackref,breaklinks,colorlinks,allcolors=cvprblue]{hyperref}
\definecolor{voencblue}{rgb}{0.69,0.87,0.94}
\definecolor{mecgreen}{rgb}{0.84,0.91,0.83}
\definecolor{mecyellow}{rgb}{1,0.95,0.8}

\newcommand{\modelname}{CineMEC} %

\renewcommand{\paragraph}[1]{\vspace{1mm}\noindent\textbf{#1}}

\newcommand{\mycolorbox}[2]{\setlength{\fboxsep}{0.4mm}\colorbox{#1}{#2}}

\newcommand{\bx}{\mathbf{x}}

\newcommand{\bz}{\mathbf{z}}

\newcommand{\mcA}{\mathcal{A}}

\newcommand{\mcM}{\mathcal{M}}
\newcommand{\mcS}{\mathcal{S}}
\newcommand{\mcF}{\mathcal{F}}
\newcommand{\mcT}{\mathcal{T}}

\newcommand{\mcR}{\mathcal{R}}
\newcommand{\mcP}{\mathcal{P}}
\newcommand{\mcB}{\mathcal{B}}

\newcommand{\mcG}{\mathcal{G}}
\newcommand{\confmark}[2][gray]{\textcolor{#1}{\scalebox{0.8}{#2}}}

\usepackage{xspace}
\makeatletter
\DeclareRobustCommand\onedot{\futurelet\@let@token\@onedot}
\def\@onedot{\ifx\@let@token.\else.\null\fi\xspace}

\def\eg{\emph{e.g}\onedot} 
\def\ie{\emph{i.e}\onedot}

\makeatother

\usepackage[capitalize]{cleveref}
\crefname{figure}{Fig.}{Figs.}
\Crefname{figure}{Figure}{Figures}
\crefname{section}{Sec.}{Secs.}
\Crefname{section}{Section}{Sections}
\crefname{table}{Tab.}{Tabs.}
\Crefname{table}{Table}{Tables}

\def\paperID{} %
\def\confName{CVPR}
\def\confYear{2026}

\title{One Identity, Many Roles: Multimodal Entity Coreference
\\ for Enhanced Video Situation Recognition}

\author{
Balaji Darur$^1$ \hspace{0.5cm}
Amanmeet Garg$^2$ \hspace{0.5cm}
Makarand Tapaswi$^1$ \\
$^1$CVIT, IIIT Hyderabad, India \hspace{0.5cm}
$^2$Amazon Prime Video, Seattle \\
{\small \url{https://katha-ai.github.io/projects/cinemec/}}
}

\begin{document}
\maketitle

\begin{abstract}
Video Situation Recognition (VidSitu) addresses the challenging problem of ``who did what to whom, with what, how, and where'' in a video.
It tests thorough video understanding by requiring identification of salient actions and associated short descriptions for event roles across multiple events.
Grounding with VidSitu requires spatio-temporal localization of key entities across shots and varied appearances.

We posit that coherent video understanding requires \emph{consistent identification} of entities that play different roles. 
We propose \emph{Multimodal Entity Coreference} (MEC) to unite entity descriptions in text with grounding across the video.
Towards this, we introduce \modelname{}, a multi-stage approach that unites event role mention groups with visual clusters of entities, without explicit grounding supervision during training.
Our approach is designed to exploit the synergy between visual grounding and captioning, where improving one influences the other and vice versa.
For evaluation, we extend the VidSitu dataset with grounding annotations.
While previous work focuses primarily on descriptions, \modelname{} improves consistency across both: captioning (+2.5\% CIDEr, +7\% LEA) \emph{and} visual grounding (+18\% HOTA).

\end{abstract}
\vspace{-3mm}
\section{Introduction}
\label{sec:intro}

\begin{figure}[t]
\centering
\vspace{-1mm}
\includegraphics[width=0.98\linewidth]{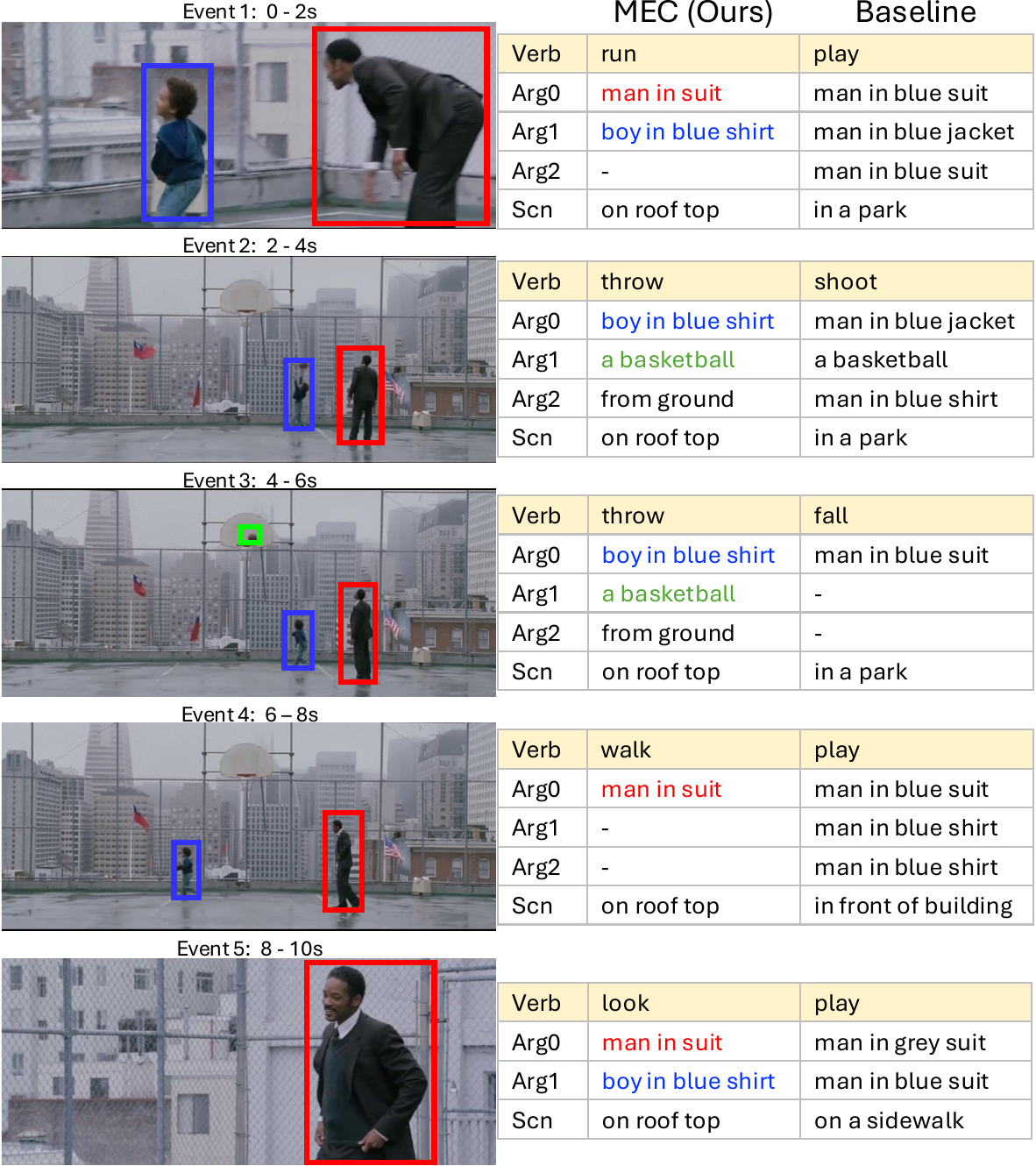}
\vspace{-2mm}
\caption{Given a video,
we present our model's outputs highlighting Multimodal Entity Coreference (MEC) in comparison to a baseline (GVSR).
Each event is tagged with the salient verb and its corresponding semantic roles (\eg~Arg0, Arg1).
MEC enforces entity-level consistency across events: the \textcolor{blue}{boy in blue shirt}, \textcolor{red}{man in suit}, and \textcolor{ForestGreen}{basketball} are tracked, linked, and described with a \textit{unique caption} across the entire video.
In contrast, the baseline produces erroneous and \textit{inconsistent} captions while referring to someone.}
\vspace{-4mm}
\label{fig:teaser}
\end{figure}

In \textit{The Pursuit of Happyness}, a scene features Chris Gardner watching his son play basketball.
Holistic understanding of this video (\cref{fig:teaser}) requires going beyond a sequence of actions such as \textit{throw} or \textit{walk}
and identifying \textit{who is throwing} (the boy), \textit{what} is being thrown (the basketball), \textit{who is watching} (the man), and \textit{where} is the action happening (on the rooftop).
Humans are naturally good at this and we also
build a narrative around entities (people and objects).
Moreover, by watching entities evolve over time, we are able to reason about their interactions and answer questions such as who is doing what and where.
However, models struggle with such compositional associations~\cite{saravanan2025velociti}.

A notable step towards the task of video semantic role labeling (SRL) is Video Situation Recognition (VidSitu)~\cite{sadhu2021vidsitu}.
Here, a video is described using a structured label space by splitting it into multiple short events, each with a salient action/verb, and tagging corresponding semantic roles (\eg~agent, patient) with descriptive arguments.
An extension, Grounded Video Situation Recognition (GVSR)~\cite{gvsr}, argues that there are multiple correct ways to describe the same entity and thus aims to ground them in the video.
Specifically, they identify a single most relevant bounding box for each SRL.
However, in both, SRLs and their grounding are treated independently across events.
They overlook the entity-centric perspective and do not connect \textit{the entity's identity with multiple roles the entity plays}, neither in SRLs, nor in the visual appearance across the video.
This results in inconsistent understanding across video events (see~\cref{fig:teaser}).

To highlight this challenge,
we formulate a new task \textbf{M}ultimodal \textbf{E}ntity \textbf{C}oreference (MEC) 
that aims to unite entity descriptions with corresponding visual tracks across the video.
We formalize MEC for VidSitu as 4 sub-tasks:
(i)~predict salient verbs for each event;
(ii)~identify entity role groups across events;
(iii)~cluster visual entity mentions (boxes) throughout the video and link them with role groups; and
(iv)~generate entity-centric SRLs.
Thus, MEC unites multimodal entity mentions (event roles, SRLs, and visual boxes) across the video.
For example, \cref{fig:teaser} shows that \modelname{} recognizes that the \textit{boy in blue shirt} takes on
the role of \textit{thrower} (agent) of the \textit{baskbetall} (thing), and later receives the attention (is \textit{looked at}) from the \textit{man in suit}.
Simultaneously, all entities are also grounded in the video with boxes that span across all frames.

\paragraph{Challenges of video MEC.} 
Recently, MEC has been studied in images~\cite{goel2023you,goel2023semi}, where rich image narratives are formed by linking entity mentions to bounding boxes.
However, the temporal dimension in videos significantly increases complexity as entities interact dynamically over time and take on different roles.
Additional visual and linguistic challenges make the problem even harder in edited videos (movies):
(i)~Visually, identifying the same entity across shots is hard as their appearance is different from varying viewpoints, may be (partially) occluded, or even absent from the view.
Addressing these requires incorporating high-level semantic reasoning about entity roles, context, and their evolution across events.
(ii)~Linguistically, the same entity may be described in multiple ways due to different shot types (\eg ``man in suit'' in a long shot or ``man with dark hair'' in a close-up).
However, for coherent understanding, it is crucial to use the same caption for an entity across events.
With multiple entities in a video (typical in movies), generating consistent \textit{and} discriminative captions to uniquely identify entities is even more crucial.
Note, these challenges are not considered by modern grounding-aware approaches~\cite{munasinghe2025videoglamm, meng2025openo3}.

\paragraph{Our approach.}
We propose \modelname{}, a four-stage approach that follows an entity-centric perspective to address MEC in VidSitu.
(i)~We cluster visual entity representations across video frames on-the-fly and use them to inform entity role representations.
(ii)~We group event roles across the video to promote consistent entity descriptions even when an entity takes on different roles across events.
(iii)~A cluster assignment module facilitates linking visual clusters with these entity role groups.
(iv)~Finally, the relevant visual cluster is fed to the captioner to encourage generation of a consistent SRL for the entity group throughout the video.

Our approach is \textit{not supervised} with ground-truth visual clustering or localization and derives its weak learning signal purely from SRLs.
As the captions are derived from visual cues of tracked entities, the approach includes a \textbf{synergistic loop}:
supervising SRL captioning improves visual clustering, and better visual clustering in turn results in improved entity role grouping and captioning.
Empirically, \modelname{} delivers substantial improvements on
\textit{captioning entities with consistency} and \textit{grounding and tracking them across shots}.

\paragraph{Contributions summary.}
(i)~We propose video Multimodal Entity Coreference that unifies the identity (textual description) of each entity that plays multiple semantic roles, with localization and tracking across the entire video.
(ii)~We propose \modelname{}, a multi-stage architecture for addressing MEC in video situation recognition.
Our approach features a synergistic improvement loop between visual clustering, entity role grouping, and captioning while being weakly supervised only with SRL captions.
(iii)~We empirically demonstrate the effectiveness of \modelname{} and see performance improvements across both vision and language metrics:
captioning (CIDEr 2.5\%),
captioning with a unique description (LEA 7\%),
localization (IoU@0.5 13\%), and
tracking (HOTA 18\%).
(iv)~We present several experiments showing ablations of each module and highlighting the synergy between entity role captioning and visual clustering.
To support evaluation, we extend the VidSitu dataset (val, test) with visual box annotations, released for future work.

\section{Related Work}
\label{sec:related}
\paragraph{Fine-grained video understanding.}
Beyond traditional coarse
action recognition~\cite{carreira2017I3d, feichtenhofer2019slowfast, girdhar2019videoactiontrans, sun2018ActorRelNet, wang2016temporal, wu2019long},
text-video retrieval~\cite{Miech2019How2100M, xu2016msr_vidret, bain2021frozen_vidret}, or
video captioning~\cite{rohrbach2017lsmdc, seo2022end_vidcap, chen2023vast_vidcap, chen2024sharegpt4video_vidcap},
fine-grained tasks require a structured or localized interpretation of actions, events, and participating entities.
They can be broadly classified as:
(i)~vision-centric challenges such as 
temporal event localization~\cite{heilbron2016fast, escorcia2016daps, shou2016temporal, zhao2017temporal}, 
spatio-temporal action detection~\cite{girdhar2019videoactiontrans, tapaswi2021ava}, and
visual tracking~\cite{yang2023trackanything, cheng2023segmentandtrackanything, cheng2022xmem, bekuzarov2023xmem++, ravi2024sam}; and
(ii)~video-language tasks such as
dense video captioning~\cite{huang2024vtimellm, yang2023vid2seq, zhou2024streaming},
video question answering~\cite{tapaswi2016movieqa, Yu2019ActivityVQA, choudhury2023zerovqa, lei2018tvqa, grunde2021agqa},
referring expression grounding and segmentation~\cite{sadhu2020video, yang2022tubedetr, liu2023grounding, kazemzadeh2014referitgame, munasinghe2023pg}.

In parallel, similar to image scene graphs~\cite{krishna2017visual},
video scene graphs have been explored to associate semantic descriptions with visual entities~\cite{luo2021moma, chen2021joint, vicol2018moviegraphs, nguyen2024hig, yang2023panoptic_pvsg, ji2020action_Action_genome}.
However, video scene-graph generation (VidSGG) often lacks support to generate free-form captions or consistently maintain entity identity across repeated instances of the same class.
Another structured and holistic task is \textit{video situation recognition} or VidSitu~\cite{sadhu2021vidsitu}, where our work lies.

\paragraph{Video situation recognition.}
VidSitu involves predicting salient actions per event and generating SRLs (or captions) of participating entities in the event roles~\cite{sadhu2021vidsitu}.
A consistent description across events is desirable to preserve entity identity and is quantitatively measured with the LEA~\cite{moosavi2016-LEA} score.
Several approaches have worked on VidSitu: 
HostSG~\cite{zhao2023hostsg}, driven from the VidSGG lens, frames the task as pairwise mapping of event roles into an entity graph;
OME~\cite{yang2023videoome} models verbs as changes in visual states of people;
ClipSitu~\cite{clipsitu} uses unpooled frame features to describe event roles;
and
TypesDev~\cite{wei2025demonstration} adopts vision-language models and retrieval-augmented generation to obtain demos during inference.

Closest to our work, GVSR (VideoWhisperer)~\cite{gvsr} introduced weakly-supervised grounding of SRLs, but restrict grounding to one box per role.
Further, role captions are predicted independently and often fail to unite occurrences of the same entity across events.
In contrast, our approach explicitly groups all entity mentions across the video and performs SRL conditioned on its corresponding visual cluster, enabling \textit{identity-aware} and \textit{entity-consistent} outputs.

\paragraph{Multimodal Entity Coreference.}
The language community has extensively studied coreference resolution~\cite{dobrovolskii2021word, bohnet2023coreference, toshniwal2021generalization} and its variants~\cite{manikantan2024major}.
Recently, it has been extended to image narratives~\cite{goel2023semi, goel2023you}, where textual mentions are linked with bounding boxes to form coherent, entity-centric descriptions.
However, extending MEC to videos is not trivial and requires tracking entities across shot changes, resolving changing entity roles, and aligning mentions with
visual tracks.
Prior efforts in identity-aware audio description (or multi-video captioning)~\cite{rohrbach2017lsmdc, park2020identity, raajesh2024micap} focused on characters and coreference, but only in text.
Differently, our approach jointly resolves textual and visual mentions across events, enabling coherent, entity-aware, and structured video understanding.

\paragraph{Grounding in MLLMs.}
Early works (\eg~VideoChatGPT~\cite{maaz2024videochatgpt}, Video-LLaMA~\cite{zhang2023videollama}) primarily focused on conversational abilities and used visual features without grounding mechanisms.
More recent advances (\eg~Qwen2.5-VL~\cite{qwen2025qwen25vl}, VideoLLaMA3~\cite{zhang2025videollama3}) demonstrate temporal understanding through dynamic frame-rate sampling or absolute time encoding and enable event localization in extended videos.

There is also notable progress in MLLMs developed for unifying visual segmentation with language~\cite{munasinghe2023pg, munasinghe2025videoglamm, peng2023kosmos, bai2024one, chen2023shikra, zhang2024llava, lee2024srtube}.
For example, pixel grounding is achieved through off-the-shelf tracker and grounding modules (PG-Video-LLaVA~\cite{munasinghe2023pg}), or extensions with dual vision encoders that emphasize spatio-temporal details (VideoGLaMM~\cite{munasinghe2025videoglamm}).
A complementary approach, Open-o3 Video~\cite{meng2025openo3} features explicit spatio-temporal evidence in text highlighting timestamps and bounding boxes to ground responses.
However, the above methods struggle with structured predictions, grounding multiple entities, and coreference across multiple text mentions and visual appearances across shots.
Our work differs in multiple ways:
(i)~we focus on entity coreference in both text and video as entities take different roles and appearances across events;
(ii)~during training, we do not need costly visual segmentation/tracking annotations as supervision is derived from SRL captions; and
(iii)~our model enables structured prediction for VidSitu.

\section{\modelname{} for Video Situation Recognition}
\label{sec:method}

\begin{figure*}[t] %
\centering
\includegraphics[width=\textwidth]{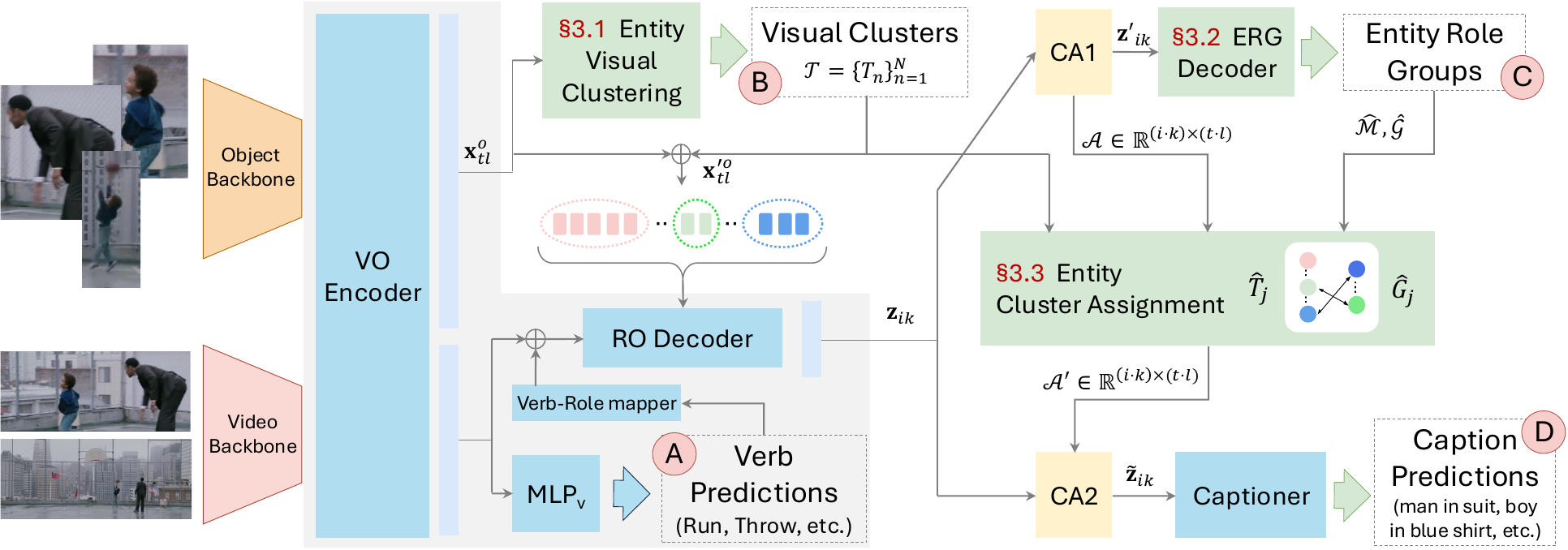}
\vspace{-5mm}
\caption{\modelname{} extends GVSR's 
\mycolorbox{voencblue}{VO encoder}, \mycolorbox{voencblue}{RO decoder}, and
\mycolorbox{voencblue}{Captioner} modules
with entity-specific modules for
\mycolorbox{mecgreen}{Visual Clustering} (EVC), 
\mycolorbox{mecgreen}{Role Grouping} (ERG), and 
\mycolorbox{mecgreen}{Cluster Assignment} (ECA).
We produce four outputs:
(A)~verb predictions and mapping to specific event-role queries,
(B)~visual clusters derived from object box proposals,
(C)~event role mention groups, and
(D)~entity-consistent captions after assigning entity role groups with the corresponding visual cluster.
\modelname{} is trained with verb, entity role group, and caption supervision, and the latter provide weak supervision for visual clustering and cluster assignment.
The architecture enables synergy between vision-language to produce coherent visual clusters and captions across multiple roles played by an entity.
}
\vspace{-4mm}
\label{fig:model_figure}
\end{figure*}

MEC for VidSitu is characterized by four interlinked tasks.
Given a video $V$ with multiple short events $\mcS = \{s_i\}$, the tasks are defined as follows.
(i)~\textit{Verb prediction} recognizes the action label $v_i$ associated with each event $s_i$.
Next, a set of roles $\mcR_i = \{ r \mid r \in \mcP(v_i) \}$ is obtained via a deterministic verb-role map $\mcP$ defined in VidSitu~\cite{sadhu2021vidsitu}.
(ii)~\textit{Entity role grouping} identifies a unique set of entities $\mcG = \{ G_j \}_{j=1}^J$ in the video by creating identity labels for each event role $\mcM = \{M_{ik}\}$.
$M_{ik}$ indexes an entity for the $k^\text{th}$ role of event $s_i$.
(iii)~\textit{Entity cluster assignment} maps each entity group $G_j$ to a visual cluster through attention scores.
The cluster is represented as a subset of boxes among all proposals $\mcB$ extracted from sub-sampled video frames to form a track.
(iv)~Finally, \textit{semantic role labeling} describes the entity group $G_j$ with $C_j$, different from the original formulation that has independent captions for each event role pair $C_{ik}$.

\paragraph{Background: GVSR~\cite{gvsr}.}
In contrast to prior works that focus mainly on verb prediction and SRL,
GVSR extends SRL with weakly supervised grounding, albeit to a single box for each event role independently.
Its architecture, VideoWhisperer, is organized around three key modules.
(i)~The \underline{Video-Object encoder (VO)} is a Transformer encoder that aligns event-level video features with object features detected across sampled frames, producing contextualized representations of both events and entities.
VO's event embeddings are used directly for verb prediction ($\hat{v}_i$).
(ii)~The \underline{Role-Object decoder (RO)} is a Transformer decoder where each event role (query) applies cross-attention to contextualized object embeddings.
Grounding is inferred by selecting the object with maximum attention, effectively linking each role to the most relevant entity in one frame.
(iii)~A \underline{Captioner}, also a Transformer decoder, generates SRL captions conditioned on outputs of the RO decoder.

While GVSR enables role-wise captioning and single-frame grounding, it does not group multiple roles of a single entity and its grounding is restricted to one frame.
This results in fragmented identities across roles and disjoint boxes for the same entity.
To overcome these limitations, we extend GVSR's architecture with three new modules (\cref{fig:model_figure}):  
(i)~Visual Clustering groups box proposals into clusters (\cref{subsec:method:vis_clus});
(ii)~Role Grouping links entity mentions across events into groups (Sec.~\ref{subsec:method:erg}); and  
(iii)~Entity Cluster Assignment assigns entity groups to visual clusters using cross-attention (Sec.~\ref{subsec:method:eca}).
These additions enable multi-role reasoning, visual grounding, and captioning, and present an entity-centric framework for holistic video understanding.

\subsection{Entity Visual Clustering (EVC)}
\label{subsec:method:vis_clus}

Tracking an entity across an entire video is challenging, particularly due to frequent shot boundaries where conventional trackers fail.
The VidSitu dataset features scenes with 2.89 shot changes on average, high for a 10 second video.
We address this challenge with a local-to-global strategy:
a tracker first links entities across frames within a shot and subsequent modules cluster tracks on-the-fly.

Similar to GVSR~\cite{gvsr}, we subsample $\mcF = \{f_t\}_{t=1}^F$ frames from the video $V$.
Each frame $f_t$ has up to $L$ proposal boxes that are linked into within-shot tracks using a prompt-free object tracker.
Together with event features, the representations of all box proposals are contextualized via the VO encoder to obtain $\bx^o_{tl}$ (for box $l$ of frame $f_t$).

Next, we link these within-shot tracks across the video by clustering $\bx^o_{tl}$ using the unsupervised FINCH algorithm~\cite{sarfraz2019efficientfinch} (see \cref{appendix:finch} for details).
This produces a set of visual clusters $\mcT = \{T_n\}_{n=1}^N$, where each cluster corresponds to a candidate entity spanning the video.
To incorporate the clustering knowledge for downstream modules, we add new \textit{cluster id embeddings} to $\bx^{o}_{tl}$, yielding cluster-aware box representations $\bx'^{o}_{tl}$.
These encourage event role queries in the RO decoder to attend to visual clusters rather than fragmented per-frame detections.

The RO decoder produces role-specific representations $\bz_{ik}$ for event $s_i$ and role $r_k$.
An additional cross-attention layer (CA1) is used to compute multimodal entity role features $\bz'_{ik}$ and the cross-attention map $\mcA \in \mathbb{R}^{(i*k) \times (t*l)}$ captures the affinity between event roles and box proposals.

\subsection{Entity Role Grouping (ERG)}
\label{subsec:method:erg}

The multimodal entity role features $\bz'_{ik}$ from CA1 contain information about roles and visual clusters, unlocking reasoning about entity identity in the video.
A key challenge is that the same entity may appear in different roles across events,
setting up the \textit{text coreference} problem.
Here, ERG's objective is to link all event role mentions that refer to the same entity.
Unlike EVC, where ground-truth visual clusters are unavailable, we use the SRL annotations to provide supervision for ERG during training%
\footnote{In VidSitu, each entity is annotated with a unique caption across roles. Thus, we derive gold mention clusters with string matching. However, text coreference approaches may also be used.}.

To group event roles, we adopt an auto-regressive formulation framed as a sequence prediction problem~\cite{raajesh2024micap}.
Our intuition is based on how humans understand a streaming video.
As we encounter new events and roles, we associate the actors with previously identified entities or create new ones.
Concretely, we train a Transformer decoder that takes the sequence $\bz'_{ik}$ as input and predicts role mention identifiers $\hat{M}_{ik}$ (a unique id for each predicted entity).
The decoder is trained sequentially with a causal mask so that future predictions do not influence the current step.
During training, we stabilize learning using teacher forcing.

During inference, the predicted mapping $\hat{\mcM}$ for each event role is converted to $\hat{\mcG}$, a set of entity role groups each mapping to the same entity:
$\hat{\mcG} = \{ \hat{G}_j \}_{j=1}^J$,
where $J$ is the number of predicted entity role groups and the group
$\hat{G}_j = \{ (i,k) \mid \hat{M}_{ik} = j \}$ is the set of event $s_i$ role $r_k$ indices mapped to the same $j^\text{th}$ entity.
Next, we use these role groups to compute entity-to-visual cluster assignment.

\subsection{Entity Cluster Assignment (ECA)}
\label{subsec:method:eca}

While ERG consolidates all entity role mentions that refer to the same entity, this grouping alone does not establish a consistent link to the visual modality.
To bridge this gap, we introduce Entity Cluster Assignment (ECA) that explicitly associates each entity role group with a single visual cluster.

ECA grounds each entity to the most relevant visual cluster using the attention scores
$\mcA \in \mathbb{R}^{(i*k) \times (t*l)}$ from CA1.
We compute this assignment by aggregating attention scores within role groups and visual clusters.
First, for each visual cluster $T_n$ (from EVC), we sum the attention scores of its constituent boxes.
Then, for each predicted entity group $\hat{G}_j$, we accumulate attention scores across its role mentions.
The combined attention map is
$\hat{\mcA} \in \mathbb{R}^{J \times N}$ and each element
$\hat{\mcA}_{jn} = \sum_{g \in \hat{G}_j} \sum_{b \in T_n} \mcA[g, b]$.
Next, for each entity group,
$\hat{T}_j = \arg\max_n \hat{\mcA}_{jn}$
is identified as the maximally attended visual cluster.
Thus, $\hat{G}_j \Leftrightarrow \hat{T}_j$ complete the MEC objective of associating each entity role group with a visual cluster.

\paragraph{Updating role embeddings for captioning.}
Given the entity-cluster assignments, we revisit the role-specific embeddings ($\bz_{ik}$) to ensure consistency during captioning.
Specifically, all role mentions linked to the same entity group $\hat{G}_j$ are constrained to attend \textit{only} to the visual boxes belonging to their assigned cluster $\hat{T}_j$.
To enforce this, we reuse parameters of the CA1 layer but replace its attention map by a fixed attention map $\mcA'$ with uniform weights for indices of mapped role groups and visual clusters and 0 elsewhere.
For simplicity, we refer to this layer as CA2.
The outputs of CA2 are \textit{visual cluster restricted entity-aware role embeddings}
$\tilde{\bz}_{ik} = \text{CA2} ( \bz_{ik}, \bx'^{o}_{tl}; \mcA' )$.
As all event roles in group $\hat{G}_j$ attend only to boxes in $\hat{T}_j$ (with equal weight), they are encouraged to learn entity consistent visual representations.

\paragraph{Captioning.}
The entity-aware embeddings $\tilde{\bz}_{ik}$ are passed to the Captioner (a Transformer decoder) to generate the SRLs.
In fact, we can generate them in two ways:
(i)~independently for each event role using $\tilde{\bz}_{ik}$, or
(ii)~for each entity group $\hat{G}_j$ by mean pooling entity-aware embeddings within the group.
During training, we observe that the model benefits from obtaining supervision from both paths, while during inference, we generate captions directly at the entity level.

Our approach yields multiple advantages:
(i)~during inference, an entity-level caption remains consistent across events and roles played by that entity;
(ii)~during training, the model is encouraged to learn similar representations for individual roles $\tilde{\bz}_{ik}$ within the group $\hat{G}_j$ and the mean pooled representation $\tilde{\bz}_j$ as they have the same target caption; and
(iii)~by controlling $\mcA'$, the captioner is forced to ground its predictions exclusively in the visual evidence of the assigned track.
By avoiding interference from other entities, we improve the quality of generated captions, and address MEC.

\subsection{Training and Inference}  

\paragraph{Training.}  
\modelname{} is trained for verb prediction, entity role grouping, and captioning in a multi-task setup.
For \underline{verb prediction}, we use the standard cross-entropy loss:
$L^{\text{v}}_i = \text{CE}(\hat{v}_i, v_i)$.
For \underline{entity role grouping}, we leverage the annotation structure in VidSitu, where an entity is assigned the exact same caption across events.
This allows us to derive the ground-truth entity role mapping $\mcM$.
Since ERG typically features a long-tail distribution, we adopt the focal loss:
$L^{\text{e}}_{ik} = \text{FL}(\hat{M}_{ik}, M_{ik})$.
Finally, we supervise \underline{captioning} in two ways:
First, a single entity \textit{group-level} caption is trained using the mean pooled embedding $\tilde{\bz}_j$ with
$L^{\text{gc}}_{j} = \sum_w \text{CE}(\hat{C}^w_{j}, C^w_{j})$, re-weighted by the number of roles in the group.
Second, individual \textit{role-level} captions are trained using $\tilde{\bz}_{ik}$ with
$L^{\text{rc}}_{ik} = \sum_w \text{CE}(\hat{C}^w_{ik}, C^w_{ik})$.
Both captioning losses are applied autoregressively over each word $w$.
Finally, the overall training objective is the sum of all losses with equal weights:
$L = L^{\text{v}} + L^{\text{e}} + L^{\text{gc}} +  L^{\text{rc}}$.

Notably, we do not supervise the entity visual clustering (EVC) or entity cluster assignment (ECA) modules, as ground-truth visual clusters with corresponding captions are unavailable.
Instead, the captioning and entity role grouping losses act as weak supervision and the model is encouraged to select clusters that align well with entities automatically as only those clusters can support generation of the appropriate caption.
In this way, captioning acts as weak supervision for both clustering and assignment.

A further advantage of our formulation is the \textit{synergy} between the linguistic and visual modalities.
Accurate role grouping in language helps the model attend to the correct visual cluster, making grounding and captioning more reliable.
Conversely, selecting the correct visual cluster reinforces role grouping, since mentions grounded to the same entity in RO are more likely to be assigned consistent entity ids.
Thus, improvements in one modality influence the other and vice versa, tightly coupling multimodal entity coreference.

\paragraph{Inference.}
At inference time, given a video $V$, we predict the salient verb $\hat{v}_i$ for each event $s_i$.
Next, we consider two options to obtain roles:
(i)~based on the ground-truth verb $v_i$, or
(ii)~based on the predicted verb via the mapper $\mcP(\hat{v}_i)$.
Following previous work, we report these as separate experiments.
In the next stage, entity visual clustering and role grouping are performed.
We rely on the predicted mention mappings $\hat{\mcM}$ to compute entity role groups $\hat{\mcG}$.
Finally, we predict a single caption per entity using the mean-pooled embeddings.
In summary, inference produces:
(i)~verb predictions $\hat{v}_i$,
(ii)~entity coreference across SRLs $\hat{G}_j$,
(iii)~association between entity role group $\hat{G}_j$ and visual cluster $\hat{T}_j$, and
(iv)~entity-consistent captions $\hat{C}_j$.

\section{Experiments}
\label{sec:experiments}

\begin{table*}[t]
\centering
\small
\tabcolsep=0.14cm
\caption{
Comparing \modelname{} against previous work on VidSitu, extended to include localization.
Metrics are grouped by subtask.
For a fair comparison, we also highlight the backbones used in each baseline:
I3D~\cite{carreira2017I3d},
SlowFast (SF)~\cite{feichtenhofer2019slowfast},
Motifs-TDE (M-TDE)~\cite{tang2020unbiased-mtde},
Faster-RCNN (FR)~\cite{ren2015faster},
YOLOE-11 tracker~\cite{wang2025yoloe} + SigLIP2~\cite{tschannen2025siglip2} (YS) features, and
BLIP2~\cite{li2023blip2}.
Trends on the test set are similar to the validation set.
}
\vspace{-3mm}
\begin{tabular}{l l l cc ccc ccc}
\toprule
& \multirow{2}{*}{Method} & \multirow{2}{*}{Backbone} & \multicolumn{2}{c}{Verb Acc.} & \multicolumn{3}{c}{SRL} & \multicolumn{3}{c}{Localization} \\
\cmidrule(r){4-5} \cmidrule(r){6-8} \cmidrule(r){9-11}
& & & @1 & @5 & CIDEr & LEA & LEA-Soft & IoU@0.3 & IoU@0.5 & HOTA \\
\midrule
\multirow{8}{*}{\rotatebox[origin=c]{90}{\textbf{Validation}}}
& VidSitu-SlowFast~\cite{sadhu2021vidsitu}~\confmark{CVPR'21} & SF & 32.64 & 69.20 & 45.52 & 50.48 & 31.99 & - & - & - \\  
& OME+OIE~\cite{yang2023videoome}~\confmark{AAAI'23} & I3D & 53.36 & 83.94 & 47.16 & - & - & - & - & - \\
& HostSG~\cite{zhao2023hostsg}~\confmark{ACMMM'23} & M-TDE & 56.15 & 86.33 & 55.09 & 55.70 & 35.01 & - & - & - \\
& ClipSitu~\cite{clipsitu}~\confmark{IJCV'25} & X-CLIP & - & - & 61.93 & 37.77 & - & - & - & - \\
& TypesDev-ucofia~\cite{wei2025demonstration}~\confmark{ICMR'25} & BLIP2 & 47.23 & - & \textbf{90.12} & 38.36 & 44.02 & - & - & - \\
& VideoWhisperer~\cite{gvsr}~\confmark{NeurIPS'22} & FR + SF & 45.06 & 75.59 & 68.23 & 48.22 & 43.93 & 41.29 & 17.43 & 11.28 \\
\rowcolor{SkyBlue!20}
& \modelname{} (Ours, old backbone) & FR + SF &  49.86 & 78.36 & {68.38} & {54.93} & {48.91} & {46.74} & {23.02} & {15.90} \\
& VideoWhisperer~\cite{gvsr}~\confmark{NeurIPS'22} & YS + SF & 46.02 & 76.49 & 73.73 & 48.37 & 48.87 & 51.49 & 42.82 & 16.23  \\  %
\rowcolor{SkyBlue!20}
& \modelname{} (Ours) & YS + SF &  49.32 & 79.83 & \textbf{76.34} & \textbf{55.78} & \textbf{52.45} & \textbf{60.15} & \textbf{55.93} & \textbf{34.22} \\
\rowcolor{gray!20}
& Human  & - & - & - & 84.85 & 72.10 & 70.33 & - & - & -  \\
\midrule
\multirow{4}{*}{\rotatebox[origin=c]{90}{\textbf{Test}}}
& VidSitu-SlowFast~\cite{sadhu2021vidsitu}~\confmark{CVPR'21} & SF & 33.94 & 70.54 & 47.25 & 50.88 & 33.50 & - & - & - \\  
& VideoWhisperer~\cite{gvsr}~\confmark{NeurIPS'22} & FR + SF & - & - & 68.04 & 48.77 & 44.57 & 40.36 & 16.89 & 10.81 \\
& VideoWhisperer~\cite{gvsr}~\confmark{NeurIPS'22} & YS + SF & - & - & 73.21 & 48.61 & 48.36 & 51.26 & 42.13 & 16.02 \\ 
\rowcolor{SkyBlue!20}
& \modelname{} (Ours) & YS + SF & - & - & \textbf{75.73} & \textbf{55.06} & \textbf{51.16} & \textbf{59.64} & \textbf{55.33} & \textbf{33.81} \\
\rowcolor{gray!20}
& Human  & - & - & - & 83.68 & 71.77 & 70.60 & - & - & -  \\
\bottomrule
\end{tabular}
\vspace{-5mm}
\label{tab:sota_result_gt_verb_roles}
\end{table*}

\paragraph{Implementation details.}
For a fair comparison to previous work, we adopt SlowFast~\cite{feichtenhofer2019slowfast} as the video backbone for $|\mcS| {=} 5$ events in the videos.
We obtain object tracks within a shot using the YOLOE-11~\cite{wang2025yoloe} prompt-free tracker.
Detections in subsampled frames are encoded using SigLIP2~\cite{tschannen2025siglip2} (with ROIAlign) as the object backbone.
Similar to GVSR~\cite{gvsr}, we sample frames at 1 fps from a 10 second video, resulting in $F {=} 11$ frames.
We cap box proposals to $L {=} 15$
for each frame, resulting in up to $|\mcB| {=} 165$ box tokens per video.
We support up to 6 roles per event, resulting in 30 event role queries, embeddings, entity maps ($\mcM$) and captions.
We also annotate entity boxes for the val and test sets, details and resulting statistics are in \cref{appendix:annotation}.

All Transformer modules%
\footnote{The VO encoder contains self-attention SA+MLP layers while the RO decoder has CA+SA+MLP layers and operates in a non-autoregressive manner.
The ERG decoder and the Captioner are autoregressive Transformers.}
have the same configurations with 3 layers, 8 attention heads, and hidden $d {=} 2048$.
The ERG decoder is set to 2 layers.
Visual clusters are obtained through two levels of FINCH~\cite{sarfraz2019efficientfinch} clustering.
The verb classifier is a single linear layer from $d{=}2048$ to 1560 verbs.
The entity id classifier in ERG is also a single linear layer from $d{=}2048$ to a maximum of 30 entity IDs.
We only consider coreference for Arg0, Arg1, Arg2, and Location/Scene roles as suggested in VidSitu~\cite{sadhu2021vidsitu}.
We use the Adam optimizer~\cite{kingma2014adam} with a learning rate of $10^{-4}$ and train the whole model end-to-end on a single L40 GPU with batch size 32.

\paragraph{Metrics.}
For verb prediction, we report Acc@K, \ie~event-level action accuracy against all ground-truth (GT) verb annotations.
For SRL we report CIDEr~\cite{Cider}, and test consistent captioning of the same entity across roles with LEA~\cite{moosavi2016-LEA} and LEA-Soft~\cite{sadhu2021vidsitu} (a combination of LEA and CIDEr per entity).
A high LEA score is obtained when the same caption is given to an entity across all its roles. %
For localization within a frame we report IoU@$\theta$ as defined by GVSR~\cite{gvsr}.
While this metric compares box predictions in a frame, we assess models' ability to cluster the entity through the entire video with a tracking metric, HOTA~\cite{luiten2021hota}.
A detailed discussion of the metrics is in \cref{appendix:metrics}.
An ideal approach:
(i)~identifies multiple roles of the same entity across the video (LEA),
(ii)~performs visual clustering and entity cluster assignment correctly (HOTA), and
(iii)~generates good captions (CIDEr).

\subsection{Comparison to State-of-the-Art} 
\label{subsec:exp:sota}
In \cref{tab:sota_result_gt_verb_roles} we compare \modelname{} against previous methods for VidSitu.
Closest to our work is the GVSR task and VideoWhisperer (VW) model~\cite{gvsr}.
Here, we reproduce VW's results with the original Faster R-CNN (FR) proposal box features, and also report results with our new YOLOE-11 tracker box proposals and SigLIP2 features (YS).
For other works, we report the scores as provided in their paper while highlighting diverse feature backbones.

Except VW, other works do not perform localization and their scores are left blank.
As VW focuses on grounding to a single box per SRL, they do not report metrics for visual grounding across the entire video (HOTA).
We compute HOTA using a post-hoc entity grouping approach:
roles with identical predicted captions are grouped into entities, and their detections are merged into visual clusters.

A recent work, TypesDev~\cite{wei2025demonstration} achieves a high CIDEr score (90.12) on VidSitu (surpassing human performance).
However, it follows a different retrieval-augmented generation pipeline to retrieve similar videos and leverage their annotations during inference for prompting MLLMs.
Nevertheless, the poor LEA score (38.36) indicates that even VLMs and LLMs struggle to describe an entity that plays multiple roles with a unique caption.
Additionally, TypesDev does not perform grounding.

\modelname{} establishes new state-of-the-art results across multiple metrics.
We improve over VW by +9 points on IoU@0.3 and +12 points on IoU@0.5, and achieve a HOTA score that is +18 points higher, owing to our role grouping and visual clustering.
For captioning, \modelname{} beats VW in CIDEr by +2.5 and LEA by a large gap +7.4, reflecting our approach's ability to capture entity coreference across SRLs.
Importantly, where prior methods show a trade-off between high CIDEr (\eg~VW~\cite{gvsr}) or high LEA (\eg~HostSG~\cite{zhao2023hostsg}), our approach achieves the best of both worlds.
LEA-Soft, a metric introduced by VidSitu, reflects this balance, and \modelname{} outperforms the best prior work (VW) by +3.5 points.
Thus, \textit{\modelname{} demonstrates strength in both tasks, explicit visual grounding of entities and implicit identity preservation through consistent, identity-aware captions.}

\begin{table}[t]
\small
\centering
\tabcolsep=0.12cm
\caption{Results for the ``predicted-verb'' setting, where roles are derived using the verb-role mapping with predicted verb $\mcP(\hat{v}_i)$.}
\vspace{-3mm}
\begin{tabular}{lcccc}
\toprule
Method & CIDEr & LEA & LEA-Soft & HOTA \\
\midrule
VidSitu~\cite{sadhu2021vidsitu} & 30.33 & 35.92 & - & - \\
TypesDev-ucofia~\cite{wei2025demonstration} & \textbf{73.71} & 30.12 & 34.97 & - \\
TypesDev-GPT4o~\cite{wei2025demonstration} & 66.56 & 27.77 & 31.89 & - \\
VideoWhisperer~\cite{gvsr} & 51.24 & 38.00 & 34.26 & \phantom{0}8.13 \\ %
\rowcolor{SkyBlue!20}
\modelname{} (Ours) & \textbf{60.32} & \textbf{46.21} & \textbf{42.64} & \textbf{24.76} \\
\bottomrule
\end{tabular}
\vspace{-3mm}
\label{tab:sota_result_gt_map_pred_verb}
\end{table}

\begin{table}[t]
\tabcolsep=0.09cm
\small
\caption{Comparing instruction-tuned MLLMs on SRL captioning against smaller models like ClipSitu and \modelname{} (Ours).
Experimental setup and results as reported in ClipSitu~\cite{clipsitu}~\confmark{IJCV'25}.}
\vspace{-3mm}
\label{tab:sota_mllms_cider}
\centering
\begin{tabular}{l ccc c c}
\toprule
Method & VILA & Qwen2-VL & LLaVA-Video & ClipSitu & \cellcolor{SkyBlue!20}\modelname{} \\
CIDEr & 40.44 & 57.28 & 60.10 & 61.93 & \cellcolor{SkyBlue!20}\textbf{76.34} \\
\bottomrule
\end{tabular}
\vspace{-3mm}
\end{table}

\begin{table}[t]
\centering
\small
\tabcolsep=0.09cm
\caption{Impact of various modules (Mod.) and losses compared to \modelname{} (row 10).
VideoWhisperer~\cite{gvsr} is in row 1 for completeness.
The evaluation metrics from left-to-right are Verb Accuracy@1, CIDEr, LEA, LEA-Soft, IoU@0.5, and HOTA.}
\vspace{-3mm}
\begin{tabular}{c l cccc cccccc}
\toprule
\# & Mod. & $L^\text{v}$ & $L^\text{rc}$ & $L^\text{gc}$ & $L^\text{e}$ & A@1 & C & L & L-S & I.5 & H \\
\midrule
1 & VW & \checkmark & \checkmark &  &  & 46.0 & 73.7 & 48.4 & 48.9 & 42.8 & 16.2  \\
\midrule
2 & \multirow{3}{*}{ERG} 
 & \checkmark & \checkmark &  & \checkmark & 47.8 & 70.1 & 52.1 & 47.8 & 50.3 & - \\
3 &  & \checkmark &  & \checkmark & \checkmark & 48.8 & 43.9 & 51.3 & 34.5 & 50.6 & - \\
4 &  & \checkmark & \checkmark & \checkmark & \checkmark & 49.0 & 70.9 & 54.1 & 49.4 & 49.4 & - \\
\midrule
5 & \multirow{3}{*}{EVC} 
 & \checkmark & \checkmark  &  &  & 50.0 & 75.9 & 50.4 & 48.5 & 54.2 & - \\
6 &  & \checkmark &  & \checkmark &  & 50.4 & 70.6 & 38.7 & 41.5 & 42.8 & - \\
7 &  & \checkmark & \checkmark & \checkmark &  & \textbf{50.7} & 76.1 & 45.0 & \textbf{47.2} & 55.5 & - \\
\midrule
8 & ERG & \checkmark & \checkmark &  & \checkmark & 48.2 & 75.6 & 55.0 & 52.1 & 53.3 & 29.1 \\
9 & + & \checkmark &  & \checkmark & \checkmark & 49.6 & 74.6 & 55.0 & \textbf{54.0} & 51.5 & 28.9 \\
10 & EVC & \checkmark & \checkmark & \checkmark & \checkmark & 49.3 & \textbf{76.3} & \textbf{55.8} & 52.5 & \textbf{55.9} & \textbf{34.2} \\
\bottomrule
\end{tabular}
\vspace{-2mm}
\label{tab:arch_loss_choice}
\end{table}

\begin{table}[t]
\centering
\small
\tabcolsep=0.09cm
\caption{Ablations for importance of applying the cross-attention after ECA with the hard coded attention map (w/o CA2) and using cluster-aware embeddings as RO's memory (no clst-emb).}
\vspace{-3mm}
\begin{tabular}{l c c c c cc c}
\toprule
Method & Acc@1 & CIDEr & LEA & LEA-S & IoU-0.5 & HOTA \\
\midrule
\rowcolor{SkyBlue!20}
\modelname{} & 49.32 & \textbf{76.34} & \textbf{55.78} & \textbf{52.45} & \textbf{55.93} & \textbf{34.22} \\
\hspace{3mm} w/o CA2      & \textbf{49.68} & 72.16 & 54.83 & 49.29 & 51.40 & 32.68 \\
\hspace{3mm} no clst-emb  & 48.37 & 74.23 & 54.17 & 51.62 & 52.36 & 33.61 \\
\bottomrule
\end{tabular}
\vspace{-5mm}
\label{tab:ca1ca2}
\end{table}

\paragraph{Predicted verb–role evaluation.}
Many prior works on VidSitu report results only on the ground-truth role setting (\cref{tab:sota_result_gt_verb_roles}).
Here, models predict the verb for each event, but the set of roles are taken from the GT verb.
While this setup simplifies the task, it does not capture end-to-end performance.
Towards holistic video understanding, we
use the predicted verb $\hat{v}_i$ for each event, and determine roles based on the verb-role mapping $\mcP(\hat{v}_i)$.
In this case, the model must predict the correct verb, and then generate entity-level captions and visual clusters for identified roles.

\cref{tab:sota_result_gt_map_pred_verb} reports results for this \textit{predicted-verb} setting.
Despite the added difficulty and missing oracle roles, our method consistently outperforms VW (and other approaches) across all metrics (+9 CIDEr, +8 LEA, +18 HOTA).
This highlights the strength of our approach to generate reliable captions and clusters in a fully end-to-end setting.

\paragraph{Comparison to captioning with MLLMs.}
\cref{tab:sota_mllms_cider} reports performance of multiple MLLMs~\cite{lin2024vila, Yao2024MiniCPMVAG, Qwen2-VL, zhang2024video} instruction-tuned for structured captioning.
We observe poor performance than a simple approach that uses CLIP features (ClipSitu~\cite{clipsitu}) and a gap of over 15 points CIDEr to \modelname{}.
A possible reason is that auto-regressive generation is not ideal as event roles are related to each other.
Instead, \modelname{} benefits from learning entity-specific representations.

\subsection{Ablations}
\label{subsec:exp:ablation}

\paragraph{Impact of modules.}  
We study the effects of excluding visual clustering (EVC) or role grouping (ERG) modules and related losses against the VW baseline in \cref{tab:arch_loss_choice}.

When using only ERG (rows 2-4), visual clusters are unavailable, and we generate captions using mean-pooled embeddings of each event-role group.
While this builds coreference across event roles, errors in role-mention grouping cause captions of one entity to be incorrectly assigned to another, resulting in a lower CIDEr and LEA score.

In contrast, when using only EVC (rows 5-7), we treat each role as a separate entity and generate captions from holistic visual clusters.
While CIDEr improves as captions are grounded in coherent visual information, LEA remains low as role mentions of the same entity are not linked.

Finally, combining both ERG and EVC (rows 8-10) provides complementary benefits.
Referencing the correct visual clusters improves role grouping, while stronger role grouping in turn improves clustering.
As a result, both CIDEr and LEA improve over the baseline.

\paragraph{Dual captioning losses}
for role and group complement each other.
Comparing row 3-4, 6-7, 9-10, we see a performance drop in CIDEr and LEA when $L^\text{rc}$ is removed.

\paragraph{Modified attention map in CA2.}
We replace the fixed attention map of CA2 with the original one from CA1.
Recall, CA2 constrains the input embedding to the captioner to boxes that are associated with the entity group.
\cref{tab:ca1ca2} ``w/o CA2'' results in weaker grounding and lower CIDEr, highlighting that entity-constrained attention is helpful.

\paragraph{Memory design in RO decoder.}
In EVC, we added cluster id embeddings to bbox representations $\bx^o_{tl}$ from the VO encoder.
Ignoring these cluster id embeddings results in a small performance drop (\cref{tab:ca1ca2} ``no clst-emb'') in CIDEr and highlights the benefit of incorporating cluster information in the RO decoder's memory representation.

\subsection{Analysis and Discussion}
\label{subsec:exp:analysis}

We refer readers to \cref{appendix:qualitative} in the supplement for qualitative analysis and example videos.
We also include some qualitative outputs from the textual grounding model Open-o3 Video~\cite{meng2025openo3} showcasing challenges of MLLMs.
Next, we present some diagnostic analyses of \modelname{} to understand how individual modules contribute to the approach.

\begin{figure}[t]
\centering
\includegraphics[width=0.8\linewidth]{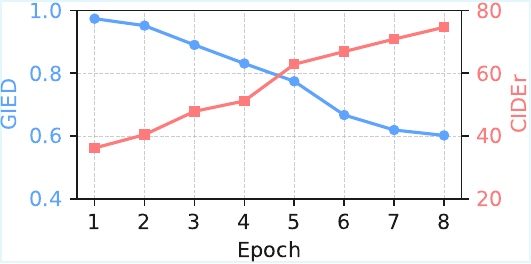}
\vspace{-3mm}
\caption{Evolution of Ground-truth Intra-Entity Distance (GIED) for visual clustering and CIDEr during training. Across epochs, GIED decreases and CIDEr increases showing the synergistic behavior of visual clustering with captioning.}
\vspace{-5mm}
\label{fig:analysis_plot}
\end{figure}

\paragraph{Visual clustering and captioning improve together.} 
We study how the captioning metric CIDEr correlates with visual clustering quality by analyzing how proposal boxes that belong to the same entity evolve during training.
Here we investigate how visual clustering improves the compactness of entity tracks.
Specifically, we measure if proposal boxes corresponding to the same entity move closer together in the embedding space as training progresses.
We compute ground-truth intra-entity distance (GIED) by associating proposal boxes with the GT entity boxes based on validation annotations with IoU $>0.3$.

To achieve this, at each frame, we match the GT entity box to the closest proposal box (within a fixed threshold), resulting in a oracle visual cluster based on proposal boxes.
Given a cluster with $n$ matched boxes, we compute the average pairwise distance across all $n\cdot(n-1)/2$ box pairs and average this across all entities to obtain GIED.
A lower value indicates that boxes belonging to the same entity cluster are coming closer in the learned representation space.

\cref{fig:analysis_plot} shows that GIED decreases consistently and correlates with an increase in CIDEr across training epochs.
This confirms the \textit{synergistic loop}: improved visual clustering benefits captioning quality and the weak learning signal of captions is able to improve clustering. 

\begin{figure}[t]
\centering
\includegraphics[width=0.8\linewidth]{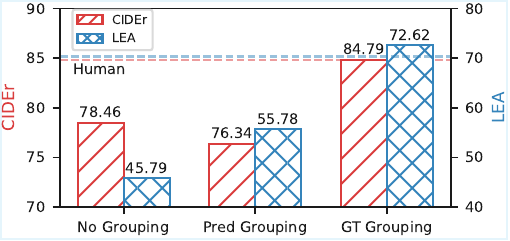}
\vspace{-3mm}
\caption{ERG strategies: no grouping, predicted grouping, and ground-truth grouping of event role mentions.
Perfect grouping yields high CIDEr and LEA, meeting human performance.}
\vspace{-6mm}
\label{fig:ana:inference_types}
\end{figure}

\paragraph{Analyzing entity role grouping.}
We evaluate the contribution of the ERG module by comparing inference without grouping or using ground-truth grouping. 
The experiments with different ERG inference strategies are performed on the same model.
As seen in \cref{fig:ana:inference_types}, we compare:
(i)~no event-role grouping, where each event-role is captioned independently;
(ii)~predicted grouping, using our ERG decoder (default \modelname{}); and
(iii)~ground-truth grouping, using the SRL string matched annotations.

Without grouping, LEA drops sharply (-10 points) as captions are generated independently for each role without any consistency across mentions of the same entity.
With predicted grouping, LEA improves, as all entity mentions are captioned consistently.
Interestingly, CIDEr decreases slightly (-2 points) due to erroneous entity id predictions in ERG resulting in incorrect grouping of some event roles.
Finally, with ground-truth grouping, both CIDEr (+8 points) and LEA (+17 points) jump significantly, indicating that our captioning module is capable of producing captions that match human-level performance of CIDEr and LEA.

ERG module's performance is limited by the long-tail nature of the role grouping problem since:
(i)~the average number of entities per video (5.2) is relatively small but the spread is high, and
(ii)~few entities (3-5) appear frequently across the video and dominate re-occurrences, also resulting in a skewed distribution.
We present additional results and discuss inference speed in \cref{appendix:quantitative}. Further limitations are discussed in \cref{appendix:limitations} in the supplement.

\vspace{-1mm}
\section{Conclusion}
\label{sec:conclusion}
\vspace{-1mm}
We proposed Multimodal Entity Coreference (MEC) as a solution for holistic understanding of videos with structured language descriptions and visual grounding.
We introduced \modelname{}, a multi-stage framework that integrates MEC into video situation recognition (VidSitu) and learns without costly visual grounding supervision.
By unifying entity role mention groups with visual clusters, \modelname{} enabled consistent identity tracking and entity-aware captions across events, advancing previous role-independent approaches.
Extensive experiments on VidSitu and comparison against task-specific works or instruction-tuned MLLMs demonstrated substantial improvements across captioning, localization, and tracking; highlighting the synergy between visual clustering and role grouping.

{\small
\noindent\textbf{Acknowledgments.}
We thank funding support from a Google India Faculty Award
and SERB SRG/2023/002544 for compute.
Amazon did not fund, direct, or influence this research work and all work was done independent of any Amazon involvement.
}

{\small
\bibliographystyle{ieee_fullname}
\bibliography{bib/longstrings, bib/references}
}

\newpage
\clearpage
\setcounter{page}{1}
\maketitlesupplementary
\appendix

In this section, we first provide a detailed discussion of the metrics (\ref{appendix:metrics}), followed by qualitative (\ref{appendix:qualitative}) and quantitative analyses (\ref{appendix:quantitative}). We then outline the limitations of our method (\ref{appendix:limitations}), describe the FINCH clustering algorithm used in the EVC module (\ref{appendix:finch}), and present our annotation pipeline and dataset statistics (\ref{appendix:annotation}).
\section{Metrics}
\label{appendix:metrics}
\paragraph{LEA}~\cite{moosavi2016-LEA} is a link-based coreference evaluation metric that weights entities by their importance and measures how well predicted mention clusters align with ground-truth clusters. The final score is computed as an F1 over precision and recall of entity links. In our setup, links are obtained by exact string matching: if two event roles share the same caption, they are treated as belonging to the same cluster.

\paragraph{LEA-Soft}~\cite{sadhu2021vidsitu} extends LEA by incorporating semantic similarity of role captions, measured using CIDEr. This ensures that even when coreference links are correct, poor or incorrect captions reduce the score.

\paragraph{IoU}~\cite{gvsr} adapts Intersection-over-Union to VidSitu grounding by computing it at the role level and averaging across all semantic roles in the video. For each role, a single predicted box is compared with the ground-truth box in the corresponding frame, and IoU@0.5 counts cases with at least 50\% overlap.

\paragraph{HOTA}~\cite{luiten2021hota}, Higher Order Tracking Accuracy, evaluates tracking by jointly considering detection, association, and localization. It balances precision and recall across trajectories, offering a unified measure of overall tracking quality.

\section{Qualitative Analysis}
\label{appendix:qualitative}
MEC in VidSitu is a challenging problem that requires identifying actions, disambiguating roles, clustering entity tracks across shot boundaries with significant visual changes, grouping event roles by their mentions, and finally generating descriptive captions for each entity. The videos, drawn from complex movie scenes with fast motion, shot changes, and diverse contexts, make this even harder. Each video contains five events. Since we obtain YOLO tracks within each shot, our Entity Visual Clustering module associates and clusters these tracks across shots, extending them to full-video entity tracks. These final tracks are paired with event-role captions and entity mentions to provide coherent role assignments throughout the video. It is important to note that the model operates only on sub-sampled frames, but the clusters can be extended throughout the video with the help of shot level tracks. Each event is represented as a table updated every two seconds. We present four representative samples in \textcolor{blue}{qualitativeresults.mp4} for full-video visualizations with tracks obtained without any explicit training
(see project website: \url{https://katha-ai.github.io/projects/cinemec/}).

\paragraph{Open-o3 Video results.}
Open-o3 Video~\cite{meng2025openo3} is a recently released open-source framework that performs grounding and object localization in videos using a large vision-language model. It associates textual queries with visual regions and outputs grounded predictions on selected frames. For qualitative comparison, we run the official Open-o3 Video inference pipeline on the same VidSitu samples used in our own qualitative analysis, following the authors' default configuration.\footnote{\url{https://github.com/marinero4972/Open-o3-Video}} For clarity of visualization, we present only the frames that Open-o3 chooses to ground, rather than every frame in the video, allowing its grounding behavior to be viewed more clearly. The resulting visualizations are provided in \textcolor{red}{openo3results.mp4}.

Despite being one of the most recent models for open-world video grounding, Open-o3 Video exhibits several limitations in the context of fine-grained video entity tracking. First, it performs grounding on only a single frame per video, preventing it from capturing entities that appear across multiple moments or scenes. Second, even within that selected frame, the model often detects only a subset of the relevant objects, leading to incomplete grounding. Third, it lacks fine-grained temporal dynamics—capabilities that are core to the VidSitu task that we target. These constraints highlight the gap that remains in current open-source grounded conversation models. Our method directly addresses these challenges by producing dense, fine-grained, and temporally continuous entity tracks across the entire video.

\section{Additional Quantitative Results}
\label{appendix:quantitative}

\begin{table}
\centering
\caption{\modelname: Results for SRL on VidSitu Validation Set. R@5: Recall at 5, C: CIDEr, R-L: ROUGE-L, C-Vb: CIDEr scores averaged across verbs,
C-Arg: CIDEr scores averaged over arguments.}
\small
\tabcolsep=0.06cm
\begin{tabular}{lccccc}
\toprule
Method & R@5 & C & R-L & C-Vb & C-Arg \\
\midrule
\textbf{VAL SET} \\
VidSitu-SlowFast~\cite{sadhu2021vidsitu}~\confmark{CVPR'21} & 23.38 & 45.52 & 42.66 & 55.47 & 42.82  \\  
OME+OIE~\cite{yang2023videoome}~\confmark{AAAI'23} & 28.72 & 47.16 & 40.86 & 53.96 & 42.78 \\
HostSG~\cite{zhao2023hostsg}~\confmark{ACMMM'23} & 29.38 & 55.09 & 43.13 & 64.24 & 47.68 \\
TypesDev~\cite{wei2025demonstration}~\confmark{ICMR'25} & 25.67 & 90.12 & 48.08 & 100.9 & 81.14\\
VideoWhisperer~\cite{gvsr}~\confmark{NeurIPS'22} & 25.25 & 73.73 & 46.21 & 82.99 & 65.52  \\ 
\rowcolor{SkyBlue!20}
\modelname{} (Ours) & 27.14 & 76.34 & 46.83 & 86.01 & 69.91 \\
\rowcolor{gray!20}
Human  & - & 84.85 & 39.77  & 91.7 & 80.15  \\
\bottomrule
\end{tabular}
\label{tab:app:more_metrics}
\end{table}

\paragraph{Multiple metrics.}
In \cref{tab:sota_result_gt_verb_roles}, we present the results of \modelname{} across multiple metrics. For brevity, only the primary metrics are reported in the main paper; however, \cref{tab:app:more_metrics} provides additional results on other commonly used metrics for the SRL task in the VidSitu benchmark. We show clear improvements over prior methods in these metrics as well.

\begin{table}[t]
\centering
\tabcolsep=0.03cm
\small
\caption{Comparision of models performance on samples from the validation set with $\geq4$ entities (n=359 of 1324).}
\begin{tabular}{lcccccc}
\toprule
Method & Acc@1 & CIDEr & LEA & L-Soft & IoU@0.5 & HOTA \\
\midrule
VideoWhisperer~\cite{gvsr} & 40.7 & 60.3 & 43.3 & 37.8 & 37.5 & \phantom{0}7.1 \\
\rowcolor{SkyBlue!20}
\modelname{} (Ours) & 44.9 & 61.5 & 46.1 & 40.1 & 51.0 & 31.7 \\
\bottomrule
\end{tabular}
\vspace{-4pt}
\label{tab:4_ents_or_more}
\vspace{-6pt}
\end{table}

\paragraph{Videos with many entities.}
Video models struggle with tracking and association of actions and entities as the number of entities increases.
To systematically analyze this effect, we construct a subset of validation videos with $\geq 4$ entities and report results in \cref{tab:4_ents_or_more}.
As expected, performance drops for all methods due to the increased complexity and long-tail nature of these samples. 
However, CineMEC consistently outperforms GVSR~\cite{gvsr} across all metrics.
This suggests that \modelname{} better enforces entity-centric understanding of videos.
Thus, while ERG is still affected by the long-tail problem, it handles such challenging scenarios more effectively than the baseline.

\paragraph{Clustering purity of ERG.}
We analyze the clustering purity of entity role groups predicted by the ERG module.
We first identify \textit{correct} and \textit{wrong} event-roles in each predicted entity cluster (purity 76\%).
When captions are generated independently (no grouping), CIDEr is (mostly) unaffected: 76.2 and 75.0.
However, with predicted grouping, CIDEr for correct roles goes up to 82.3 \ie~close to human performance (83.7) and that for wrong roles reduces to 67.8.
This causes the overall CIDEr to reduce by 2 points compared to no-grouping in \cref{fig:ana:inference_types}.

\paragraph{Compute cost.} Inference time for CineMEC is only \SI{0.65}{\second} per video on 1$\times$ A6000 GPU.
FINCH is a fast algorithm with $\mathcal{O}(N \log N)$ complexity~[53] and running it on the fly is fast.
The preprocessing time to extract Yolo + Siglip2, SlowFast visual features is \SI{3.7}{\second} per video. Our model has 245M parameters as compared to GVSR's 220M.

\section{Limitations}
\label{appendix:limitations}

\paragraph{Semantic Role Labeling.}
\paragraph{Entity Role Grouping.}
The long-tail distribution also affects grouping: entities appearing early in a video or more frequently across events dominate ID assignments, while rare entities are clustered wrongly.
This makes grouping harder and constrains downstream tasks. As shown in our second analysis, improving grouping accuracy could substantially benefit both coreference and captioning.

\paragraph{Visual Clustering.}
The object proposal boxes occasionally miss relevant entities, leading to incomplete clusters and weaker grounding and captioning. This limitation arises not just from our framework but also from the underlying object proposal model.

\section{FINCH-based Clustering}
\label{appendix:finch}

\paragraph{Input.}
We begin by describing how the box representations used for clustering are obtained. For each shot in the video, we apply a prompt-free tracker (YOLOE~\cite{wang2025yoloe}) to generate short per-shot tracklets. From these tracklets, we collect the proposals corresponding to our sub-sampled frames. Each proposal is then passed through the VO encoder, yielding contextualised object features $\bx'^{o}_{tl}$ for box $l$ of frame $f_t$.

\paragraph{Overview of FINCH.}
FINCH~\cite{sarfraz2019efficientfinch} is a hierarchical clustering algorithm that forms partitions purely from first-order nearest-neighbor relations. At every level, each point is linked to its closest neighbor, and clusters naturally emerge without the need for specifying the number of clusters or setting a distance threshold. The algorithm operates on a pairwise distance matrix that captures how dissimilar the feature representations of any two points are, and uses this matrix to determine nearest-neighbor links. In our case, FINCH clusters the set of contextualised box features $\bx'^{o}_{t,l}$ into long, video-level entity tracks, which requires constructing such a distance matrix over these features.

\paragraph{Why the Standard FINCH Matrix Requires Adaptation.}
Directly applying FINCH assumes that all points are independent, which is violated in video-based tracking. Our boxes exhibit two domain-specific constraints that, if ignored, lead to invalid cluster assignments. We therefore modify the FINCH distance matrix to respect these constraints.

\subparagraph{Constraint 1: Boxes from the same frame cannot belong to the same cluster.}
In frame \(t\), multiple proposals \(\bx'^{o}_{t,l}\) may appear simultaneously, but a valid temporal track may contain \emph{only one} box per frame. Standard FINCH may incorrectly merge same-frame proposals into a single cluster. We enforce a hard exclusion by setting
\begin{equation}
\mathcal{D}\!\left(\bx'^{o}_{tl},\, \bx'^{o}_{tl'}\right) = \infty 
\quad \text{for all } l \neq l'.
\end{equation}
thereby preventing any clustering between proposals originating from the same frame.

\subparagraph{Constraint 2: Boxes within the same shot-level tracklet must remain together.}
The tracker provides reliable identity continuity within each shot. Thus, all proposals belonging to the same shot-level tracklet correspond to the same underlying entity. Standard FINCH does not utilize this structural cue and may split such proposals across clusters.
For any feature pair drawn from the same shot-tracklet, we strongly encourage early merging by scaling their distances:
\begin{equation}
\mathcal{D}\!\left(\bx'^{o}_{tl},\, \bx'^{o}_{t'l'}\right)
\;\leftarrow\;
10^{-5}\,\mathcal{D}\!\left(\bx'^{o}_{tl},\, \bx'^{o}_{t'l'}\right).
\end{equation} where \text{for all } (t,l),(t',l') in the same shot-level tracklet. This ensures that shot-consistent proposals cluster together before FINCH proceeds to merge across different shots. 

With these modifications, FINCH produces clusters that extend shot-level tracklets into coherent, video-level entity trajectories. We employ a two-level hierarchy to achieve this: the first level groups boxes into stable intra-shot tracks while preventing same-frame overlaps, and the second level merges these shot-level groups across the full video to form long-range entity tracks.

\paragraph{Summary.}  
Our modified FINCH pipeline (i) enforces frame-wise constraints and (ii) clusters tracks within shots. Together, these steps yield continuous visual clusters across shots, tightly integrating clustering with downstream video-language grounding.

\section{Annotation Pipeline}
\label{appendix:annotation}

\begin{figure}
\centering
\includegraphics[trim=0 7cm 0 0, clip, width=1\linewidth]{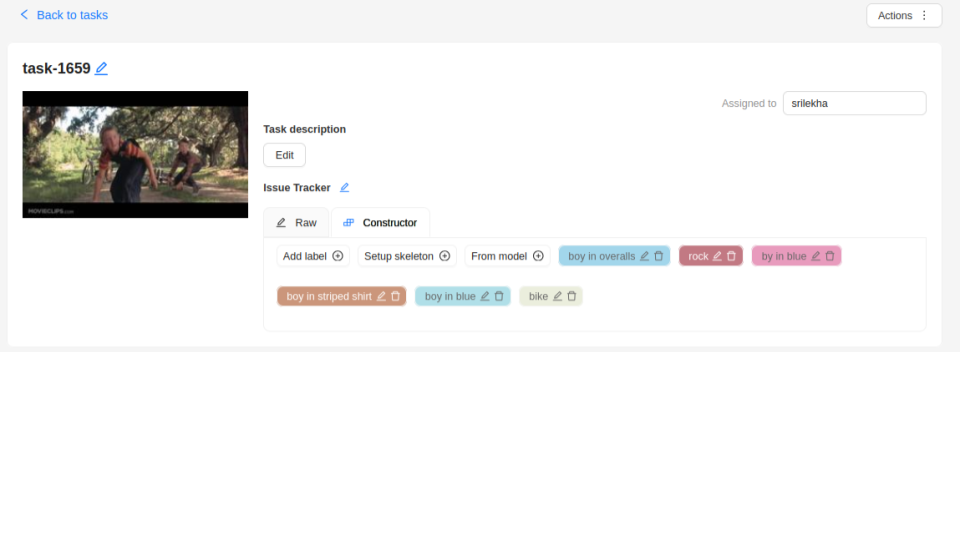}
\caption{Example annotation task for a video. There are a total of 11 frames sub-sampled at T=1 second from a 10 second video. Text highlighted in colors represent different labels.}
\label{fig:supp:taskfigure}
\end{figure}

\textbf{Annotation Platform Setup.}  
For a 10-second video containing 5 events, we sample frames at 1 fps, $\mcF = \{f_t\}_{t=1}^F$, resulting in 11 frames, following the protocol in GVSR~\cite{gvsr}. From the SRL annotations, we collect captions for the primarily visual roles \textit{agent (Arg0), patient (Arg1), and instrument (Arg2)} across all events. The unique captions from these roles are retained and treated as ground-truth labels for the video $V$. Each video is then used to create a dedicated annotation task in the CVAT tool~\cite{cvat}, with video-specific labels as illustrated in \cref{fig:supp:taskfigure}. We annotated bounding boxes for videos in the validation and test splits of the VidSitu dataset, with overall statistics reported in \cref{supp:table:annot_stats_table}.

\begin{table}[t]
\caption{Statistics of our grounding annotations on the VidSitu dataset (together for validation and test sets).}
\centering
\begin{tabular}{l r}
\toprule
\textbf{Statistic} & \textbf{Value} \\
\midrule
Number of videos & 2810 \\
Number of unique captions & 8157 \\
Total boxes annotated & 48026 \\
Avg. boxes per unique caption & 5.89 \\
Avg. boxes per video & 17.09 \\
\bottomrule
\end{tabular}
\label{supp:table:annot_stats_table}
\end{table}

\begin{figure}
\centering
\includegraphics[width=1\linewidth]{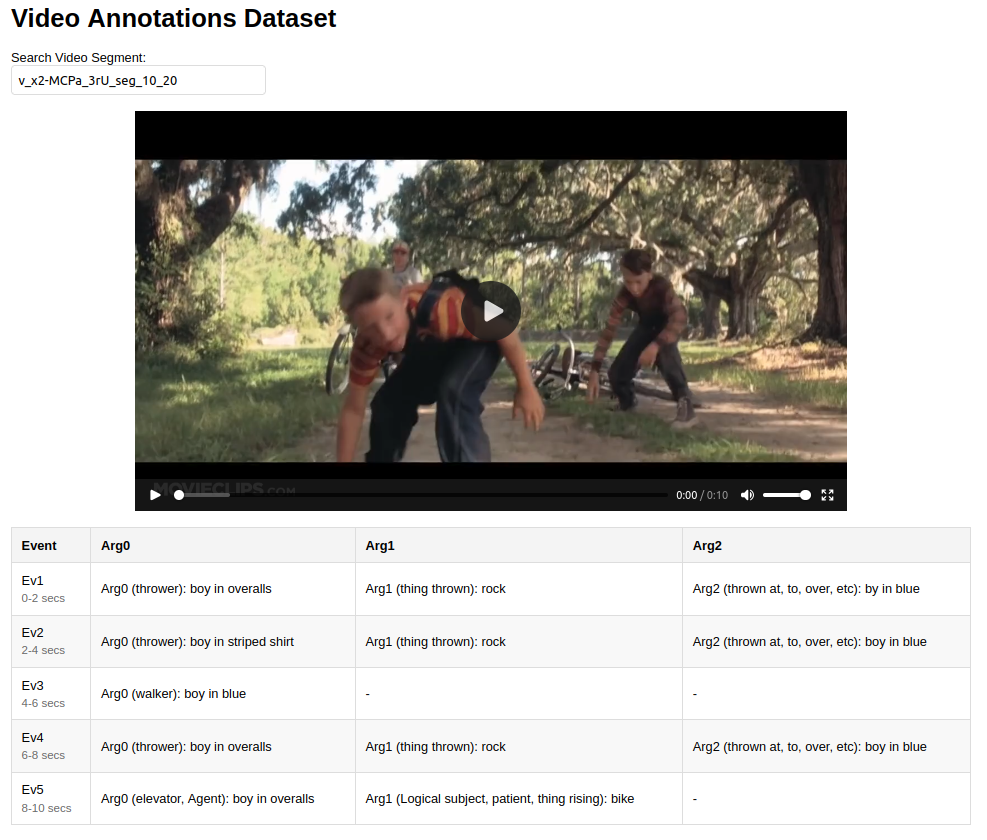}
\caption{Example visualization of ground truth semantic role labels for each event of a video. Annotators first go through the video and identify associations between captions and visual entities accurately.}
\label{supp:fig:video_vis_forest_gump}
\end{figure}

\subsection{Annotation Process}
\begin{figure}
\centering
\includegraphics[width=1\linewidth]{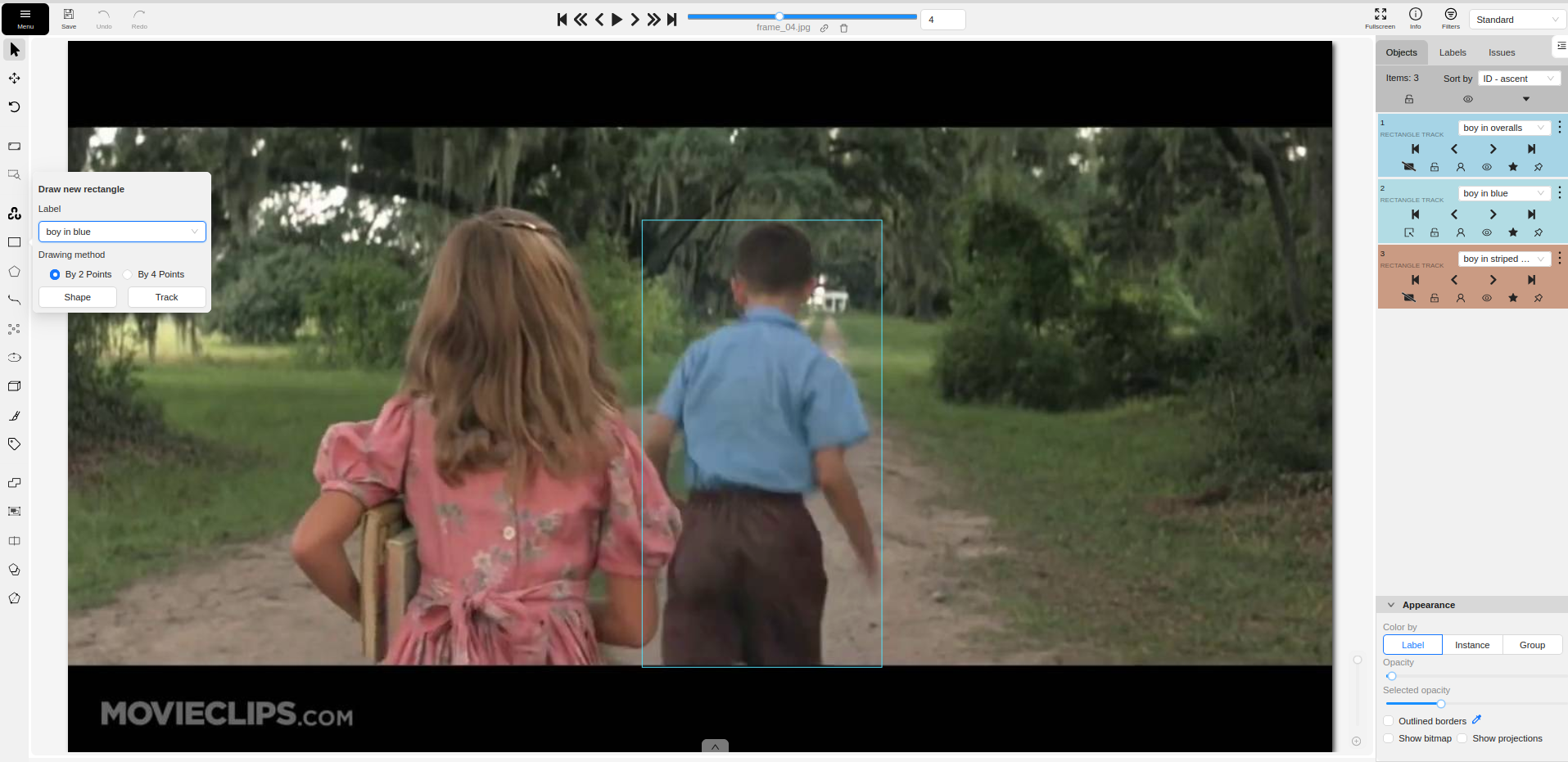}
\caption{Choose a label that can be visually identified, then draw a bounding box around it or extend the existing track of the recognized entity. Label \textit{boy in blue} is visible in \textit{frame 05.}}
\label{supp:fig:frame5_forest_gump}
\end{figure}
\begin{figure}
\centering
\includegraphics[width=1\linewidth]{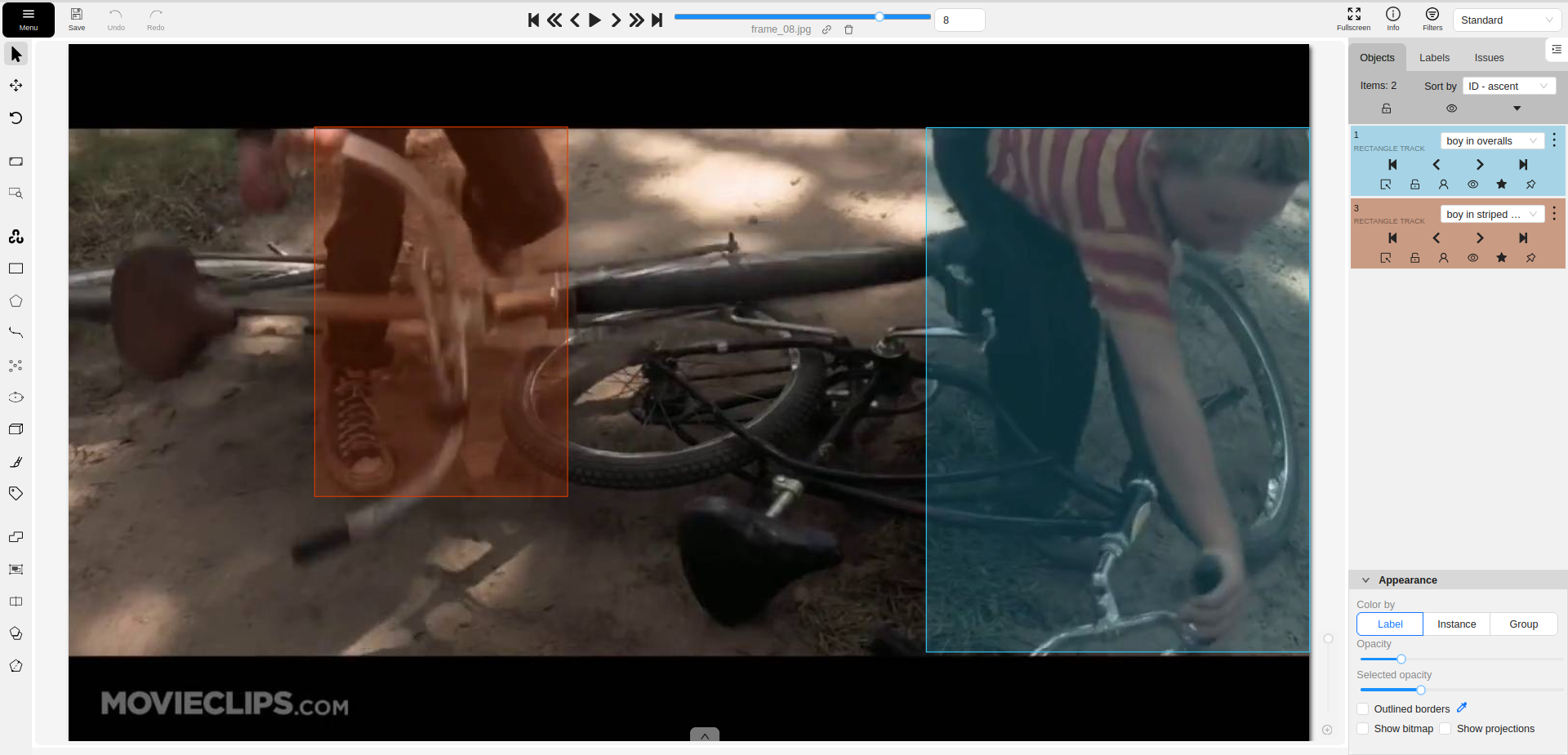}
\caption{Identifying and linking entities to their captions becomes particularly challenging during shot changes like the one above. In such cases, association requires more than a simple object match; it demands a compositional understanding of semantic cues and spatial positioning within the scene to correctly distinguish between entities and assign the appropriate caption.}
\label{supp:fig:frame8_forest_gump}
\end{figure}
The annotation process begins with annotators first watching the complete video along with its associated role captions\cref{supp:fig:video_vis_forest_gump}. This initial pass provides a global understanding of the scene and helps identify entities and their continuity across shot changes. Annotators then return to the CVAT tool~\cite{cvat} to assign each caption to its corresponding entity by drawing bounding boxes in the sampled 11 frames (see \cref{supp:fig:frame5_forest_gump}, \ref{supp:fig:frame8_forest_gump}).  

\begin{figure}
\centering
\includegraphics[width=1\linewidth]{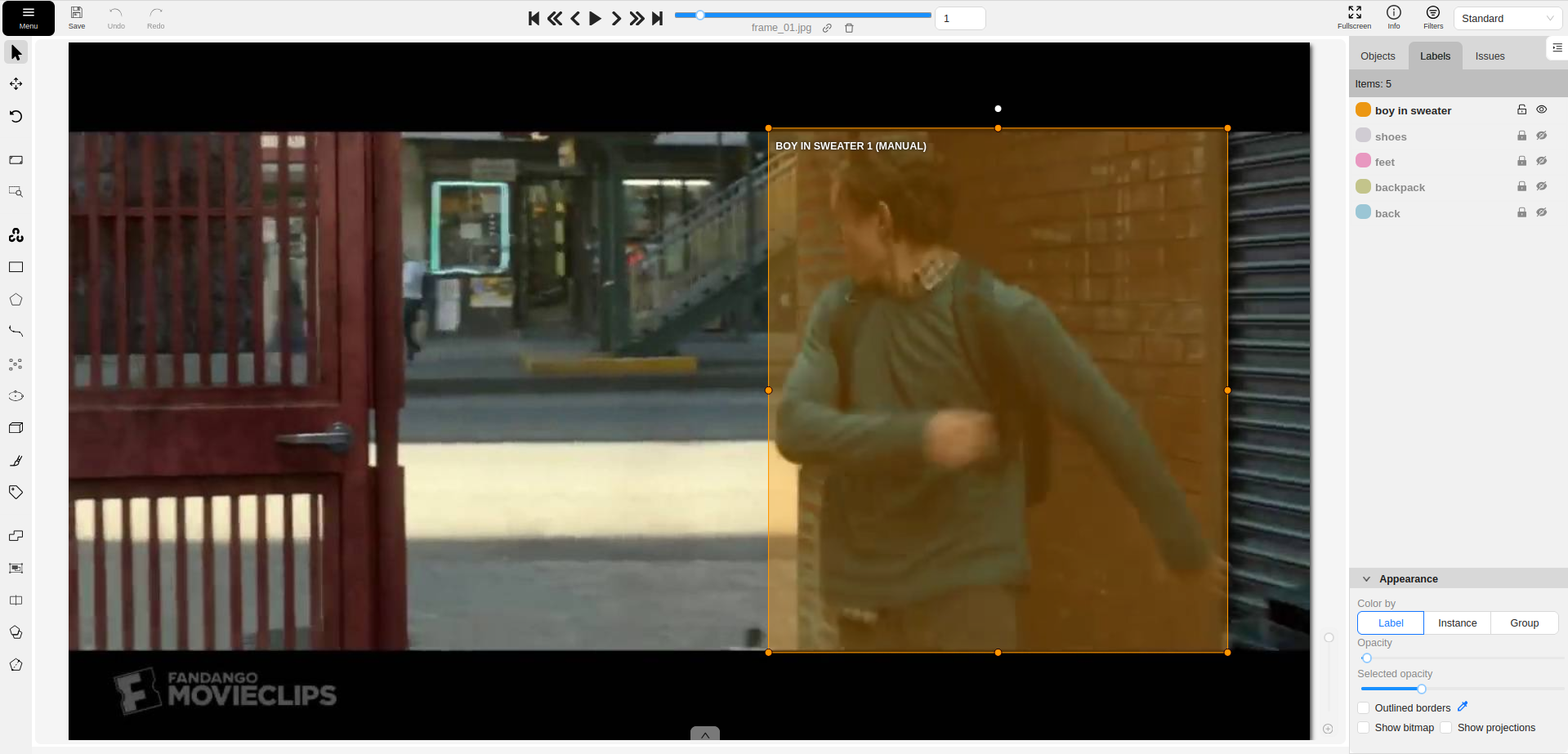}
\caption{Label \textit{back} is a non-visual role, hence it is not grounded.}
\label{supp:fig:back_not_groundable}
\end{figure}

While most captions can be localized visually, certain cases are inherently non-annotatable, such as abstract expressions (\eg, \textit{back} in \cref{supp:fig:back_not_groundable}). In other cases, annotation is non-trivial due to complex shot transitions. For example, in the \textit{Forest Gump} sequence, multiple shot changes require viewing the full video to ensure consistent tracking of people. Similarly, \cref{supp:fig:frame8_forest_gump} highlights a particularly challenging case, where multiple boys with bicycles must be distinguished and grounded. Here, visual cues such as dress color and spatial positioning, rather than faces, enable correct identification. These examples illustrate how annotating VidSitu requires not only bounding-box labeling but also a deeper semantic understanding of context and continuity. We assess potential annotator bias and observe a 92\% inter-annotator agreement, measured using pairwise F1 (IoU@0.5) on 10 sampled videos, indicating strong consistency and minimal bias across annotations.

\paragraph{}

\paragraph{Compensation.}
We fairly compensated the annotators for their efforts.

\end{document}